\documentclass[a4paper, 10pt, conference]{ieeeconf}

\IEEEoverridecommandlockouts
\usepackage[utf8]{inputenc}
\usepackage[noadjust]{cite}
\usepackage{amssymb}
\usepackage{amsmath}
\usepackage{bm}
\usepackage{verbatim}
\usepackage{upgreek}
\usepackage{svg}
\DeclareMathOperator*{\argmax}{arg\,max}

\usepackage{graphicx}
\usepackage{multirow}
\usepackage{caption, subcaption}
\usepackage{tikz}
\usepackage{tikz-3dplot}
\usetikzlibrary{bayesnet}
\usepackage[ruled, linesnumbered]{algorithm2e}
\usepackage{lmodern}
\usepackage{url} 
\usepackage{mathtools}

\newcommand{\coordinatesystem}[1]{$\underrightarrow{\mathcal{F}}_{\text{#1}}$}

\title{\LARGE \bf
Implicit representation priors meet Riemannian geometry \\ for Bayesian robotic grasping
}

\author{Norman Marlier$^{1*}$ 
Julien Gustin$^{2}$ %
Olivier Brüls$^{3}$ %
Gilles Louppe$^{4}$ %
\thanks{*The authors come from the University of Liège, Belgium}
\thanks{$^{1}${\tt\small norman.marlier@uliege.be}}%
}

\begin{document}

\maketitle
\thispagestyle{empty}
\pagestyle{empty}

\begin{abstract}

Robotic grasping in highly noisy environments presents complex challenges, especially with limited prior knowledge about the scene. In particular, identifying good grasping poses with Bayesian inference becomes difficult due to two reasons: i) generating data from uninformative priors proves to be inefficient, and ii) the posterior often entails a complex distribution defined on a Riemannian manifold. In this study, we explore the use of implicit representations to construct scene-dependent priors, thereby enabling the application of efficient simulation-based Bayesian inference algorithms for determining successful grasp poses in unstructured environments. Results from both simulation and physical benchmarks showcase the high success rate and promising potential of this approach.

\end{abstract}

\section{Introduction}


Grasping is a fundamental skill for any robotic system. While current methods are effective for highly constrained tasks in structured environments, new and complex applications require increased flexibility and more advanced algorithms to account for the uncertainties that emerge in unstructured and noisy environments. Bayesian inference offers a well-principled approach to address these uncertainties; however, robotic tasks present unique challenges that make Bayesian inference difficult to apply, particularly for sampling-based algorithms. Firstly, many Bayesian approaches assume that the likelihood is tractable and can be evaluated, which is seldom the case in robotics. Secondly, parameters may span a vast space, leading to inefficient sampling strategies. Lastly, parameters of interest often belong to smooth Riemannian manifolds, further complicating the inference procedure.
In this paper, we address these challenges by designing informative scene-dependent priors and using simulation-based inference algorithms combined with geometric sampling methods. Our contributions are summarized as follows:
\begin{itemize}
\item We integrate simulation-based Bayesian inference methods~\cite{cranmer2020frontier} with 3D implicit representations for robotic grasping.
\item We adapt geodesic Monte Carlo~\cite{byrne2013geodesic} with a neural ratio estimator to sample on Riemannian manifolds with an intractable likelihood.
\item We validate our method through simulated and real experiments, demonstrating promising grasping performance.
\end{itemize}

\section{Problem statement}

\begin{figure}
\centering
    \resizebox{\linewidth}{!}{
        \includegraphics[height=5cm]{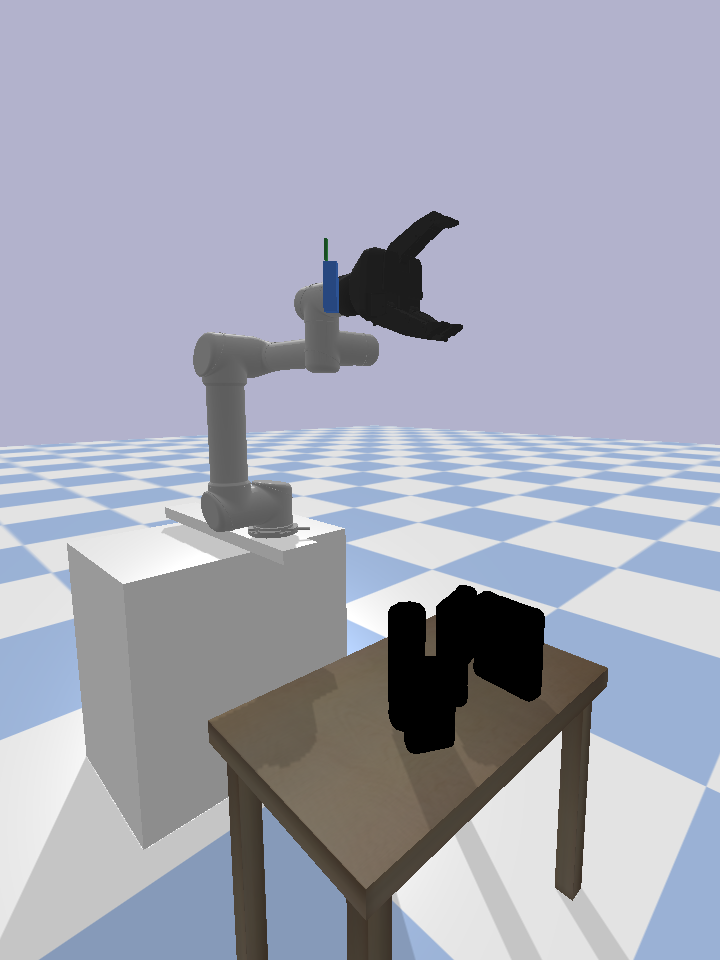}
        \enspace
        \includegraphics[height=5cm]{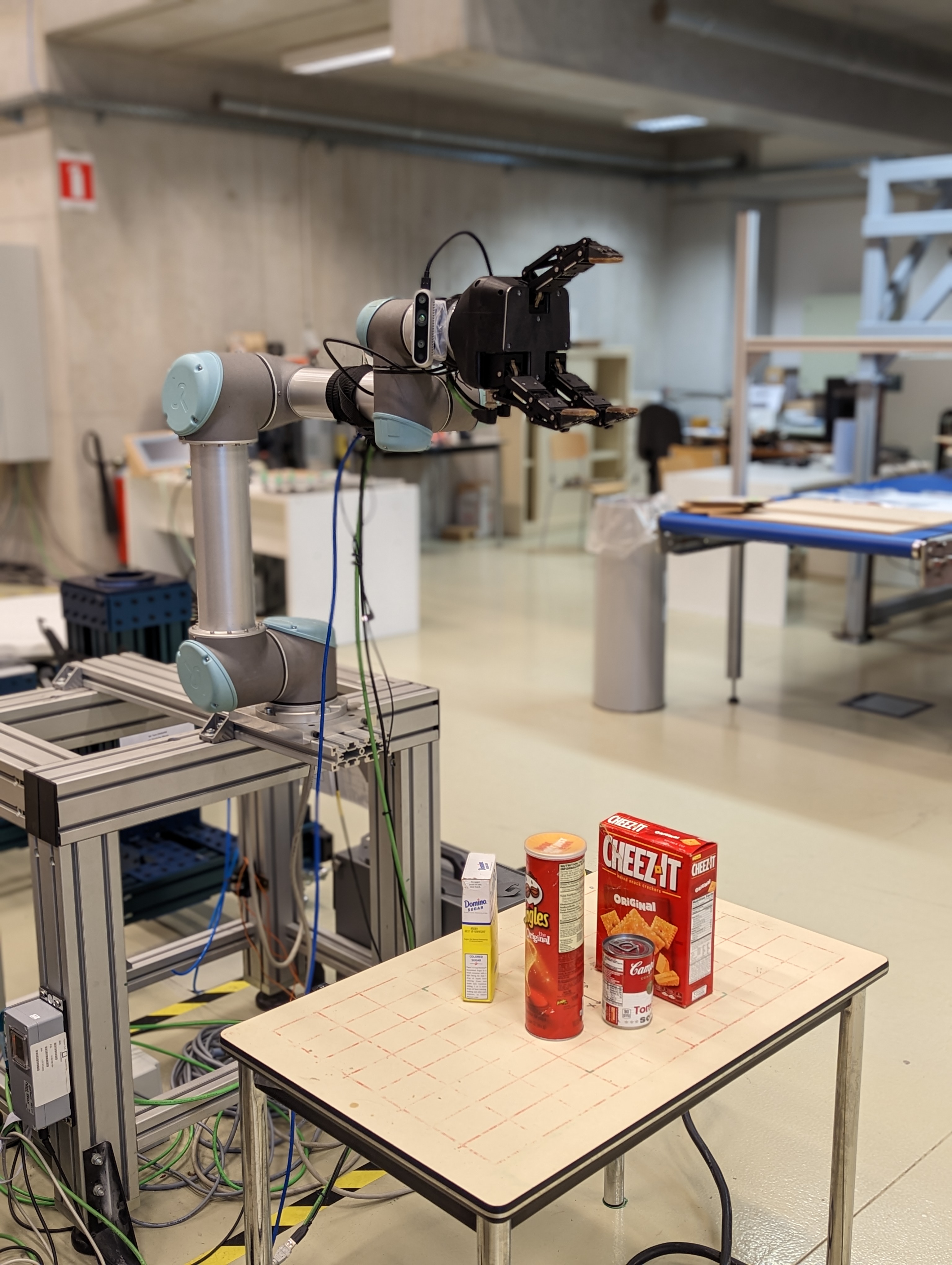}
    }
    \caption{Our benchmark scene. (left) The simulated environment. (right) The real setup.}
    \label{fig:scene}

\end{figure}

We consider the problem of planning 4-DoF hand configurations for a robotic gripper handling various unknown objects on a table, observed with a depth camera. A benchmark scene is shown in Fig.~\ref{fig:scene}.

\subsection{Notations}
\textbf{Frames} We use several reference frames in our work. The world frame \coordinatesystem{W} and the workspace frame \coordinatesystem{S} can be chosen freely and are not tied to a physical location. The world frame is used for the robot and the sensor, while the workspace frame is used for our inference system. \coordinatesystem{C} and \coordinatesystem{E} correspond respectively to the camera and the tool centre point.
\par
\textbf{Hand configuration} The hand configuration $\mathbf{h} \in \mathcal{H} = \mathbb{R}^{3}\times \mathbb{S}^{1}$ is defined as the pose $(\mathbf{x},\mathbf{q}) \in \mathbb{R}^{3}\times \mathbb{S}^{1}$ of the hand, where $\mathbf{x}$ is the vector $\Vec{SE}$ expressed in \coordinatesystem{S} and $\mathbf{q}$ is the planar rotation represented with complex numbers defined in \coordinatesystem{S}.
\par
\textbf{Binary metric} A binary variable $S \in \{0, 1\}$ indicates if the grasp fails ($S=0$) or succeeds ($S=1$).
\par
\textbf{Observation} Given the depth image $I$ with its corresponding transformation camera to world $\mathbf{T}_{\text{WC}}$ and camera intrinsic matrix $K$, we construct a point cloud $\mathbf{P} \in \mathbb{R}^{2048\times3}$ expressed in \coordinatesystem{S}.
\par
\textbf{Occupancy} A binary variable $o \in \{0, 1\}$ indicates if a point $\mathbf{p} \in \mathbb{R}^{3}$ is occupied by any object of the scene.
\par
\textbf{Latent variables} Unobserved variables $\mathbf{z}$ capture uncertainties about the nonsmooth dynamics of contact, the sensor noise, as well as the number of objects and their geometry.

\subsection{Grasping as inference}
We formulate the problem of grasping as the Bayesian inference of the hand configuration $\mathbf{h}^{*}$ that is a posteriori the most likely given a successful grasp, an occupancy $o$ and a point cloud $\mathbf{P}$. That is, we are seeking the maximum a posteriori (MAP) estimate
\begin{equation}
\label{eq:map}
\mathbf{h}^{*} = \argmax_{\mathbf{h}}~p(\mathbf{h}|S=1, o=1, \mathbf{P}),
\end{equation}
from which we then compute the joint trajectory
\begin{equation}
    \tau_{1:m} = \Lambda(\tau_{0}, \textsc{ik}(\mathbf{h}^{*}), \mathbf{P}),
\end{equation}
where $\textsc{ik}$ is an inverse kinematic solver, $\tau_{1:m}$ are waypoints in the joint space, $\tau_{m} = \textsc{IK}(\mathbf{h}^{*})$ with $\mathbf{h}^{*}$ expressed in \coordinatesystem{W} and $\Lambda$ is a path planner.

\section{Implicit representation of priors and posteriors for robotic grasping}
\label{sec:method}

From the Bayes rule, the posterior of the hand configuration is
\begin{equation}
\label{eq:posterior}
p(\mathbf{h}|S, o, \mathbf{P})  = \frac{p(S \mid \mathbf{h}, o,  \mathbf{P})}{p(S\mid o, \mathbf{P})}p(\mathbf{h}\mid o, \mathbf{P}), 
\end{equation}
which can be rewritten as the product of the likelihood-to-evidence ratio $r$ and a scene-dependent prior
\begin{equation}
\label{eq:posterior_ratio}
    p(\mathbf{h}|S, o, \mathbf{P}) = r(S\mid \mathbf{h}, o, \mathbf{P})p(\mathbf{h}\mid o, \mathbf{P}).
\end{equation}
\subsection{Priors}
\label{prior:position}

\textbf{Position} The scene-dependent prior over the position $\mathbf{x}$ is the distribution $p(\mathbf{x} | o, \mathbf{P})= \frac{p(o | \mathbf{x}, \mathbf{P})}{p(o | \mathbf{P})} p(\mathbf{x})$, where $p(o | \mathbf{x}, \mathbf{P})$ is the likelihood of the occupancy $o$, $p(\mathbf{x})$ is uniform over the workspace, and $p(\mathbf{x} | \mathbf{P})$ is simplified to $p(\mathbf{x})$ by independence.

We model the occupancy likelihood $p(o | \mathbf{x}, \mathbf{P})$ using a Convolutional Occupancy Network~\cite{peng2020convolutional}. This network computes the occupancy by first producing three canonical features planes $\mathbf{c}_{xy}(\mathbf{P}), \mathbf{c}_{xz}(\mathbf{P})$ and $ \mathbf{c}_{yz}(\mathbf{P})$. Then, the bilinear interpolations of the three planes are used to compute $\psi(\mathbf{P}, \mathbf{x}) = \mathbf{c}_{xy}(\mathbf{P})(\mathbf{x}) + \mathbf{c}_{xz}(\mathbf{P})( \mathbf{x}) + \mathbf{c}_{yz}(\mathbf{P})(\mathbf{x})$. These point-wise features at point $\mathbf{x}$ are finally processed by a fully connected network, outputting the occupancy probability.

This implicit representation  allows us to sample interesting grasping positions from $p(\mathbf{x} | o, \mathbf{P}) \propto p(o | \mathbf{x}, \mathbf{P}) p(\mathbf{x})$. We use Hamiltonian Monte Carlo (HMC) to take advantage of the differentiability of the occupancy network.

\begin{figure*}[ht]
    \centering
    \resizebox{\linewidth}{!}{
    \includegraphics{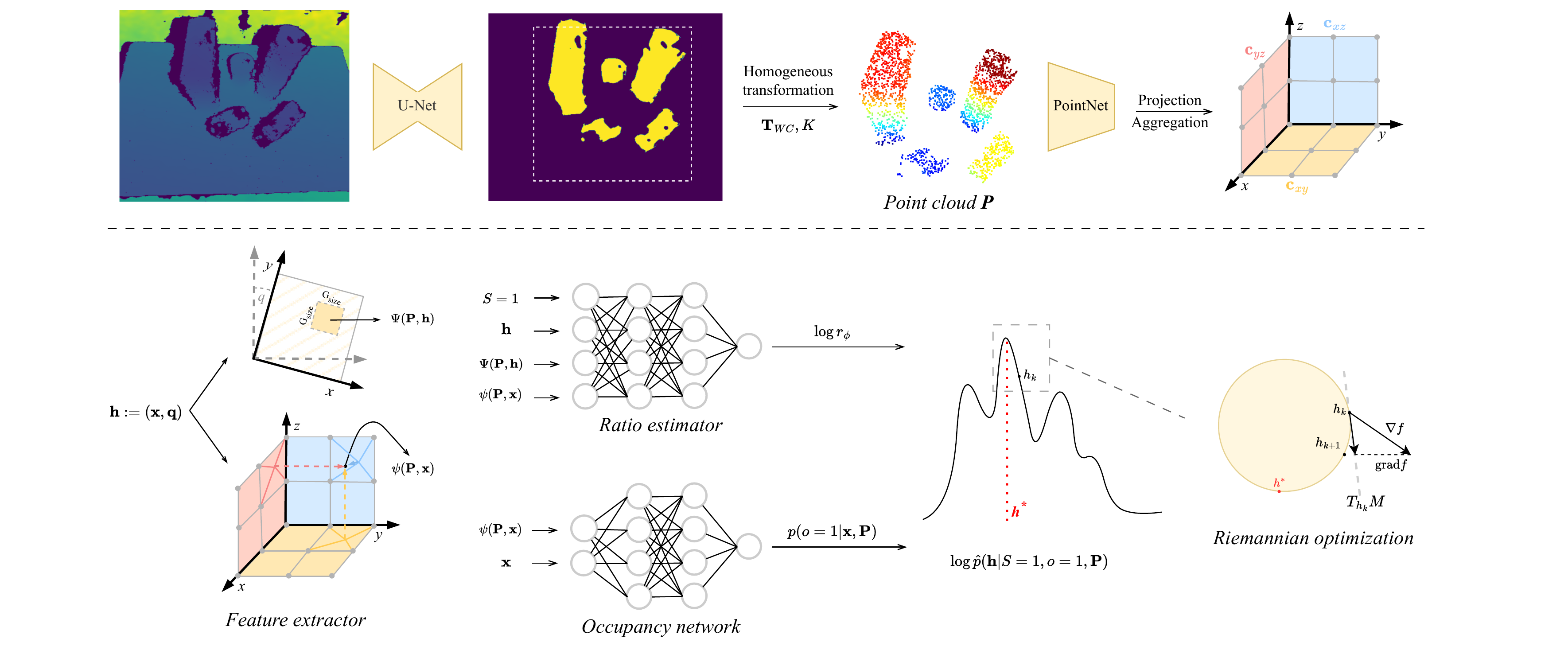}
    }
    \caption{Our grasp inference pipeline, as run on the scene of Fig.~\ref{fig:scene}. It begins with a noisy depth image of the scene, from which we first separate the objects from the background using a U-Net~\cite{ronneberger2015unet}. We then generate three canonical feature planes following the approach in~\cite{peng2020convolutional}. To evaluate a given $\mathbf{h}$, we extract point-wise $\psi(\mathbf{P}, \mathbf{x})$ and local $\Psi(\mathbf{P}, \mathbf{h})$ features and feed them to the ratio and occupancy networks. Using the resulting differentiable posterior and Riemannian optimization, we finally identify the most plausible hand configuration $\mathbf{h}^{*}$.}
    \label{fig:pipeline}
\end{figure*}

\par
\textbf{Orientation} The prior of the orientation $\mathbf{q}$ is a uniform distribution over the unit circle $\mathbb{S}^{1}$. This prior is \textit{invariant} to any rotation $\mathbf{R} \in \mathrm{SO(2)}$ applied to $\mathbf{q}$, satisfying $p(\mathbf{q})=p(\mathbf{R}\mathbf{q})$. This property enables free selection of the reference frame on the table. Additionally, the prior can be extended to $\mathrm{SO(3)}$ by using quaternions on $\mathbb{S}^{3}$.

\par
\textbf{Hand configuration} Finally, the prior of the hand configuration is $p(\mathbf{h}\mid o, \mathbf{P}) = p(\mathbf{x}\mid o, \mathbf{P}) p(\mathbf{q})$.

\subsection{Ratio}
The likelihood function $p(S\mid\mathbf{h}, o, \mathbf{P})$ and the evidence $p(S\mid o, \mathbf{P})$ are both intractable. However, drawing samples from forward models remains feasible with physical simulators, hence enabling likelihood-free Bayesian inference algorithms. In particular, the likelihood-to-evidence ratio $r(S\mid \mathbf{h}, o, \mathbf{P})$ (Eq.~\eqref{eq:posterior_ratio}) can be approximated by a neural network $r_{\phi}(S\mid \mathbf{h}, o, \mathbf{P})$ using amortized neural ratio estimation~\cite{pmlr-v119-hermans20a}. Here, instead of using only point-wise features $\psi(\mathbf{P}, \mathbf{x})$ as input for the ratio, we add a crop of the features plane $\mathbf{c}_{xy}(\mathbf{P})$, centred at $\mathbf{x}$, sized by the gripper and rotated by $\mathbf{q}$. These features $\Psi(\mathbf{P}, \mathbf{h})$ local in the neighbourhood of the grasping point are nearly equivariant to a 2D transformation applied to the object, \textit{i.e} $\textbf{T}\Psi(\mathbf{P}, \mathbf{h}) \approx \Psi(\textbf{\text{T}}\mathbf{P}, \mathbf{h}), \textbf{T} \in \text{SE}(2)$.

\subsection{Posteriors}
Given our scene-dependent prior and our likelihood-to-evidence ratio, we approximate the posterior over the hand configurations as
\begin{equation}
    \hat{p}(\mathbf{h}\mid S, o, \mathbf{P}) = r_{\phi}(S\mid \mathbf{h}, o, \mathbf{P})p(\mathbf{h}\mid o, \mathbf{P}).
\end{equation}

This approximation defines an implicit function~\cite{murphy2021implicit} on the product of manifolds $\mathbb{R}^{3}\times \mathbb{S}^{1}$ that is both fully tractable and differentiable, allowing the use of gradient-based methods for computing the MAP and sampling. Therefore, we can use Markov Chain Monte Carlo methods to sample from our posterior approximation $\hat{p}(\mathbf{h}\mid S, o, \mathbf{P})$.
In particular, based on \cite{pmlr-v119-hermans20a}, we use a likelihood-free version of HMC by replacing the intractable likelihood with the ratio. The potential energy function is defined as $U(\mathbf{h}) \triangleq -\log p(S\mid \mathbf{h}, o, \mathbf{P})$ and its difference is $U(\mathbf{h}_{t}) - U(\mathbf{h}') = \log r(S \mid \mathbf{h}_{t}, \mathbf{h}')$. The gradient used in the integration step is given by $\nabla_{\mathbf{h}}U(\mathbf{h}) = -\nabla_{\mathbf{h}}\log r(S\mid \mathbf{h}, o, \mathbf{P})$. 
To account for the geometry of the parameter space, we then further extend our likelihood-free HMC with a geodesic integrator. The geodesic Monte Carlo scheme uses geodesic flow to perform the integration while staying on the manifold. To this end, orthogonal projections and geodesics are needed in a closed form.
Finally, geodesic Monte Carlo can be applied to a product of manifolds $\mathcal{M}_{1} \times \mathcal{M}_{2} : \{(x_{1}, x_{2}): x_{1}\in \mathcal{M}_{1}, x_{2}\in \mathcal{M}_{2}\}$, such as $\mathbb{R}^{3}\times \mathbb{S}^{1}$ in our specific case. Geodesic flow can be executed in parallel; only the evaluation of the ratio requires both variables $\mathbf{x}$ and $\mathbf{q}$.
In this manner, we can sample from the posterior density defined on a smooth manifold with closed-form geodesic.
The full sampling procedure is summarized in Algorithm~\ref{alg:gHMCNRE} of Appendix~\ref{appendix:algorithm}.

\section{Experiments}
We assess our approach on a robotic grasping task in both simulation and real-world settings. We generate data in a \textit{packed} scenario, as defined in \cite{breyer2020volumetric}. Additional experiments can be found in Appendix~\ref{appendix:sampling_orientation} and~\ref{appendix:sampling_position}.

\subsection{Grasp inference pipeline}
Starting from the depth image $I$, we remove the background and extract only pixels of the objects with a segmentation model based on a U-Net architecture~\cite{ronneberger2015unet}. Then, we convert $I$ to a point cloud $\mathbf{P}$ with $2048$ points, which passes through an encoder and produces three canonical feature planes. We then extract point-wise $\psi(\mathbf{P}, \mathbf{x})$ and local $\Psi(\mathbf{P}, \mathbf{h})$ features to evaluate the occupancy and the ratio networks. To smooth the posterior approximation, we use an ensemble of 6 ratio models. Finally, we compute the MAP by maximizing the log posterior density~\cite{marlier2023simulation}. To this end, we use a Riemannian gradient ascent which preserves the nonlinearity of $\mathbb{S}^{1}$. 
A visual summary of our method is given in Fig.~\ref{fig:pipeline}.

\begin{figure}
    \centering
    \resizebox{\linewidth}{!}{
    \includegraphics{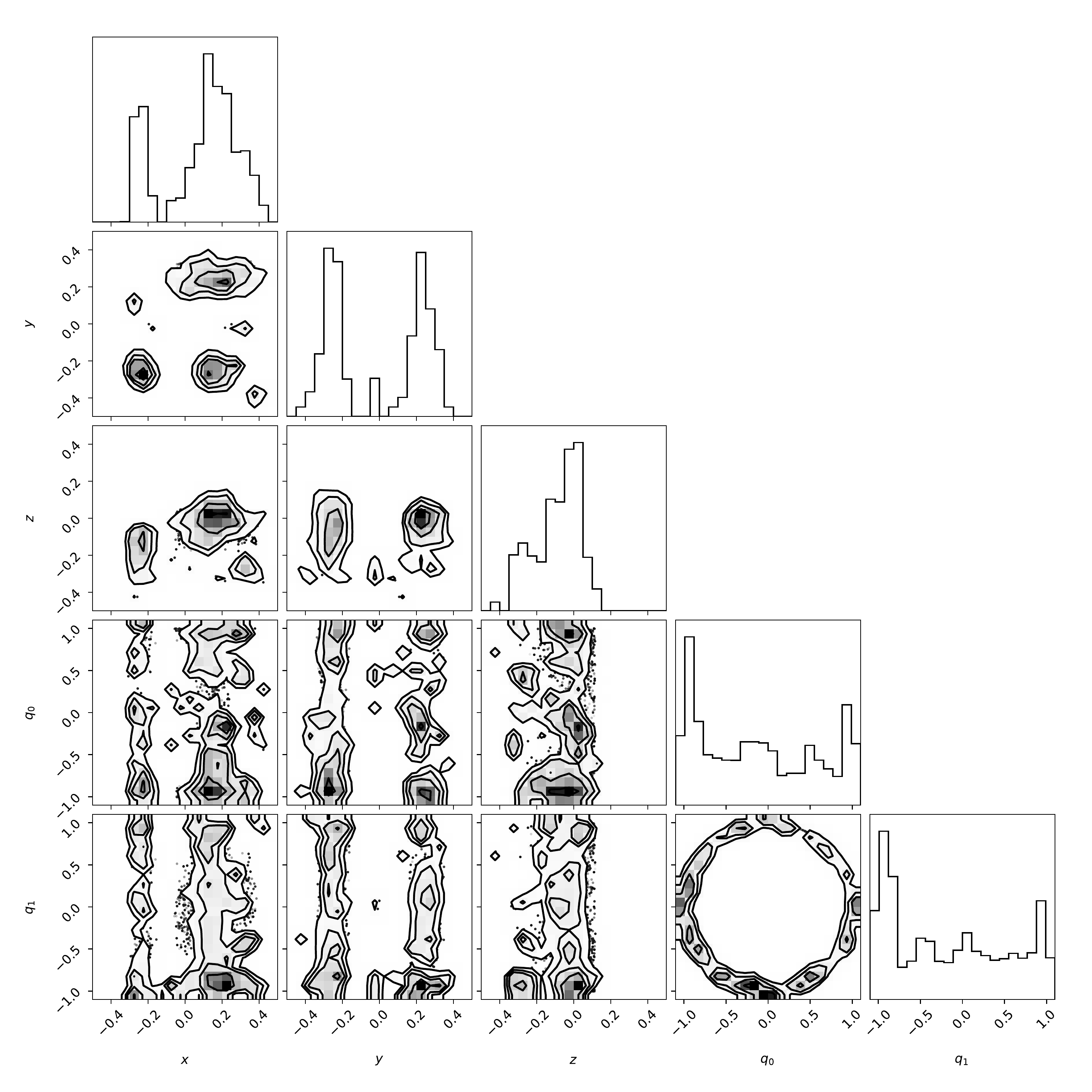}}
    \caption{Estimated posterior distribution $\hat{p}(\mathbf{h}\mid S=1, o=1, \mathbf{P})$ of plausible successful hand configurations, for the scene shown in Fig.~\ref{fig:scene}.}
    \label{fig:gmhc}
\end{figure}
\begin{figure}
\centering
    \resizebox{\linewidth}{!}{
        \includegraphics[height=5cm]{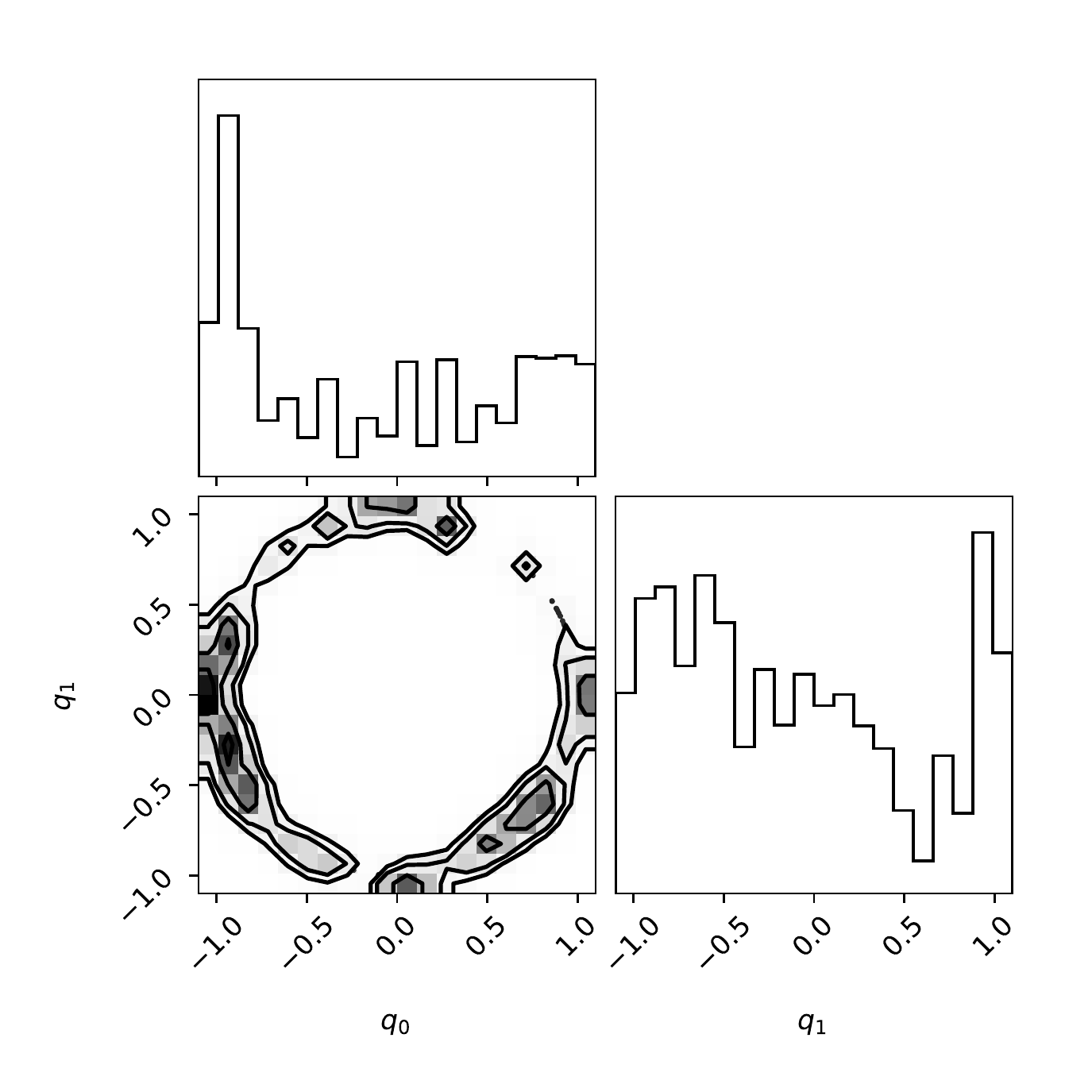}
        \enspace
        \includegraphics[height=5cm]{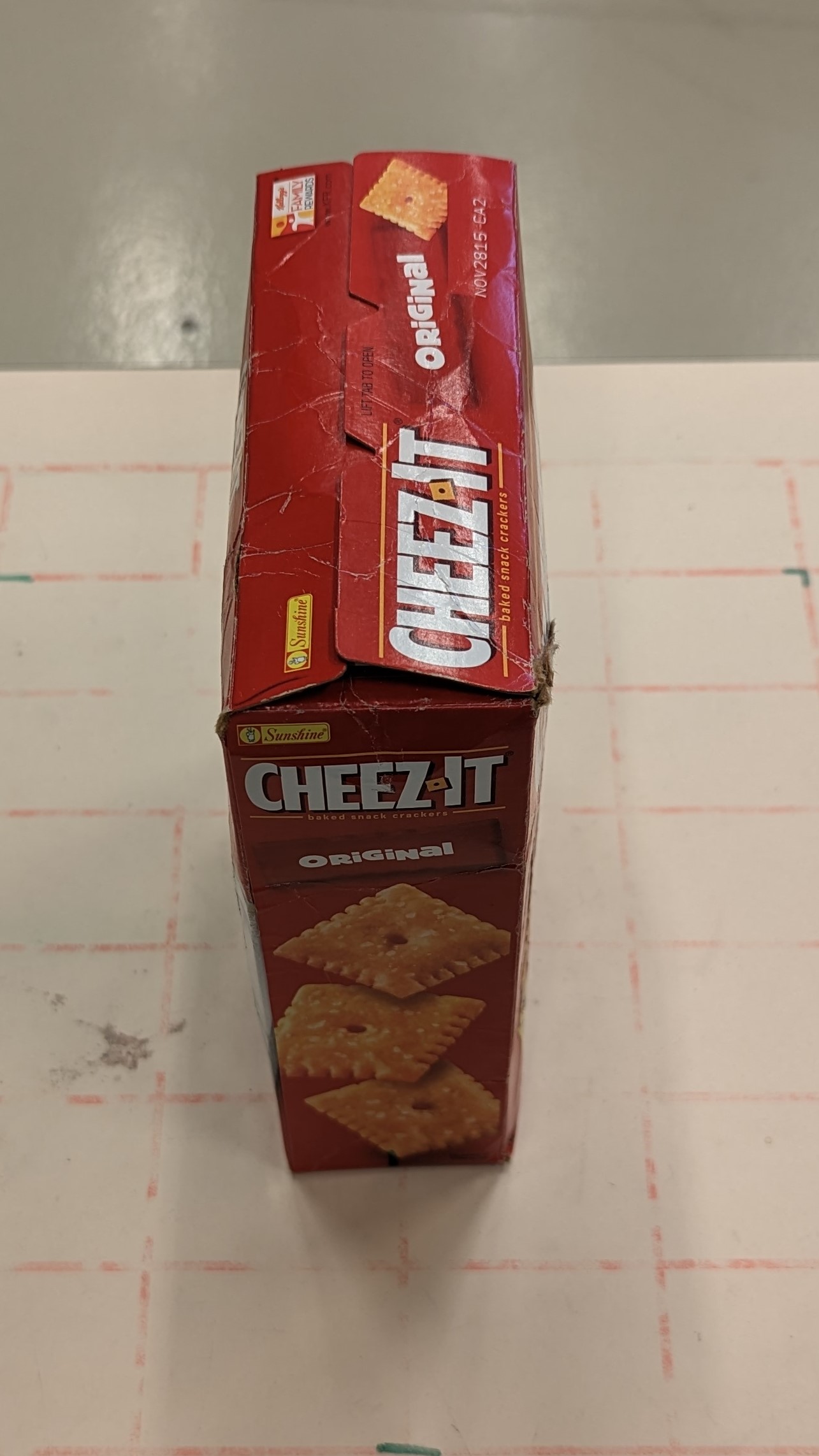}
    }
\caption{(left) Estimated posterior distribution of the orientation $\hat{p}(\mathbf{q}\mid S=1, o=1, \mathbf{P})$. (right) Single object. }
    \label{fig:ghmc_ori}
\end{figure}

Our likelihood-free geodesic Monte Carlo is used to sample plausible hand configurations $\mathbf{h} \sim \hat{p}(\mathbf{h}\mid S=1, o=1, \mathbf{P})$ for successful grasps, as shown in Fig.~\ref{fig:gmhc}. Although our conditional prior distributes density across everywhere on the objects, the posterior assigns minimal density to the bottom of objects when multiple objects are present on the table. This occurs due to potential collisions between the gripper and the table, or the gripper and other objects.
Regarding rotation, the posterior resembles the prior because multiple objects on the table, some with axial symmetry, allow for a wide range of grasp orientations. When only a single object is used, distinct modes can be observed, indicating that our posterior captures meaningful orientations, as shown in Fig.~\ref{fig:ghmc_ori}.


\subsection{Simulation results}

To compare our approach, we evaluated it against Grasp Pose Detection (GPD) \cite{ten2017grasp} and  Volumetric Grasping Network (VGN) \cite{breyer2020volumetric} in terms of success rate and percent cleared \cite{breyer2020volumetric}  using the same dataset and a similar scenario (Table~\ref{tab:results}). Our model achieved a high success rate of 91.1\%, which is very close to VGN's best success rate of 91.5\%. However, our model operated in a more constrained setup with 4 DoF instead of 6, limiting the gripper's movement and increasing the risk of collisions. This limitation resulted in a slightly lower percent cleared compared to VGN. Furthermore, the gripper used was primarily designed for larger objects, making it challenging to handle smaller ones.

\subsection{Real-world results}

\begin{figure}
\centering
    \resizebox{\linewidth}{!}{
        \includegraphics[height=2cm]{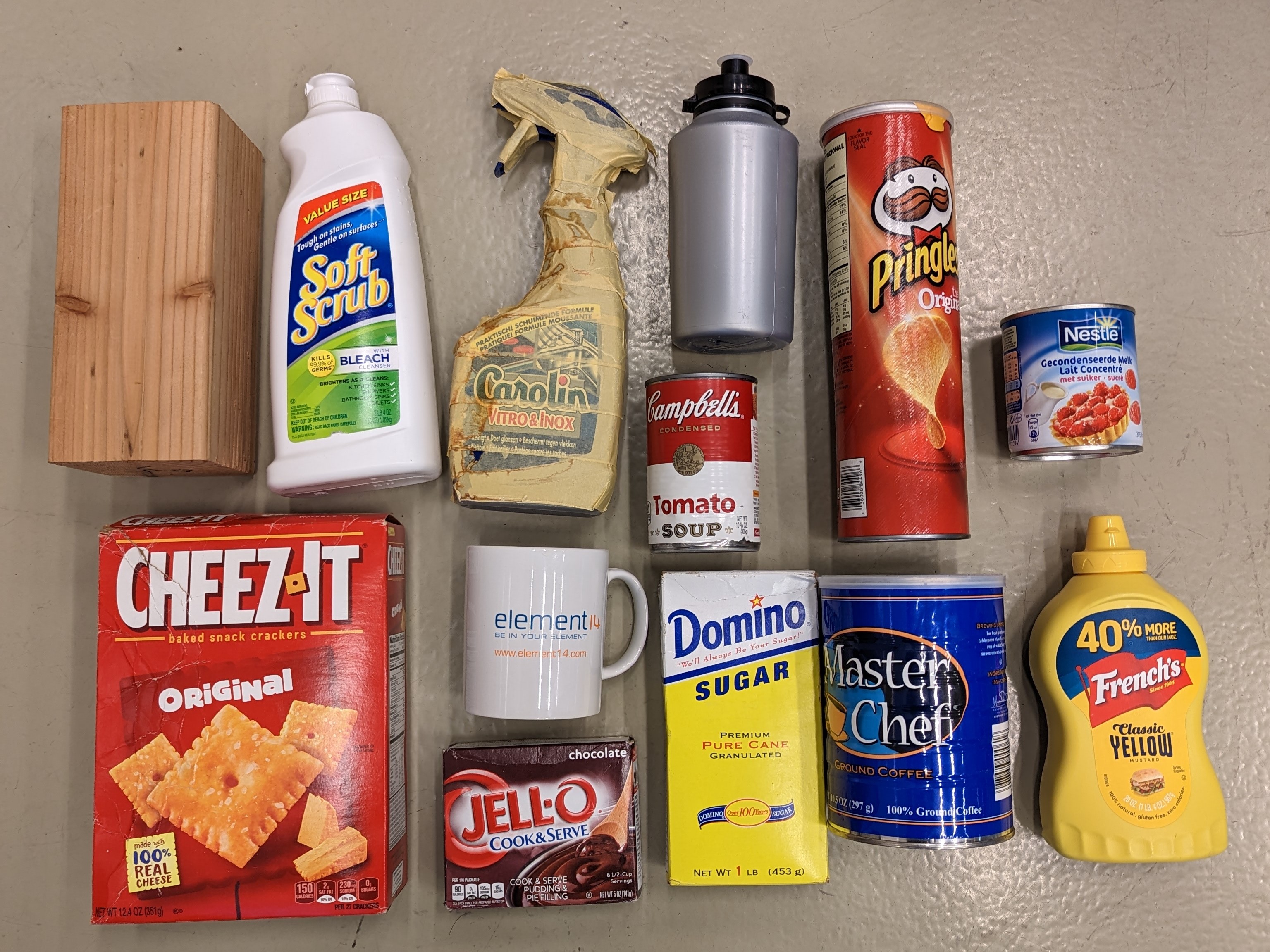}
        \enspace
        \includegraphics[height=2cm]{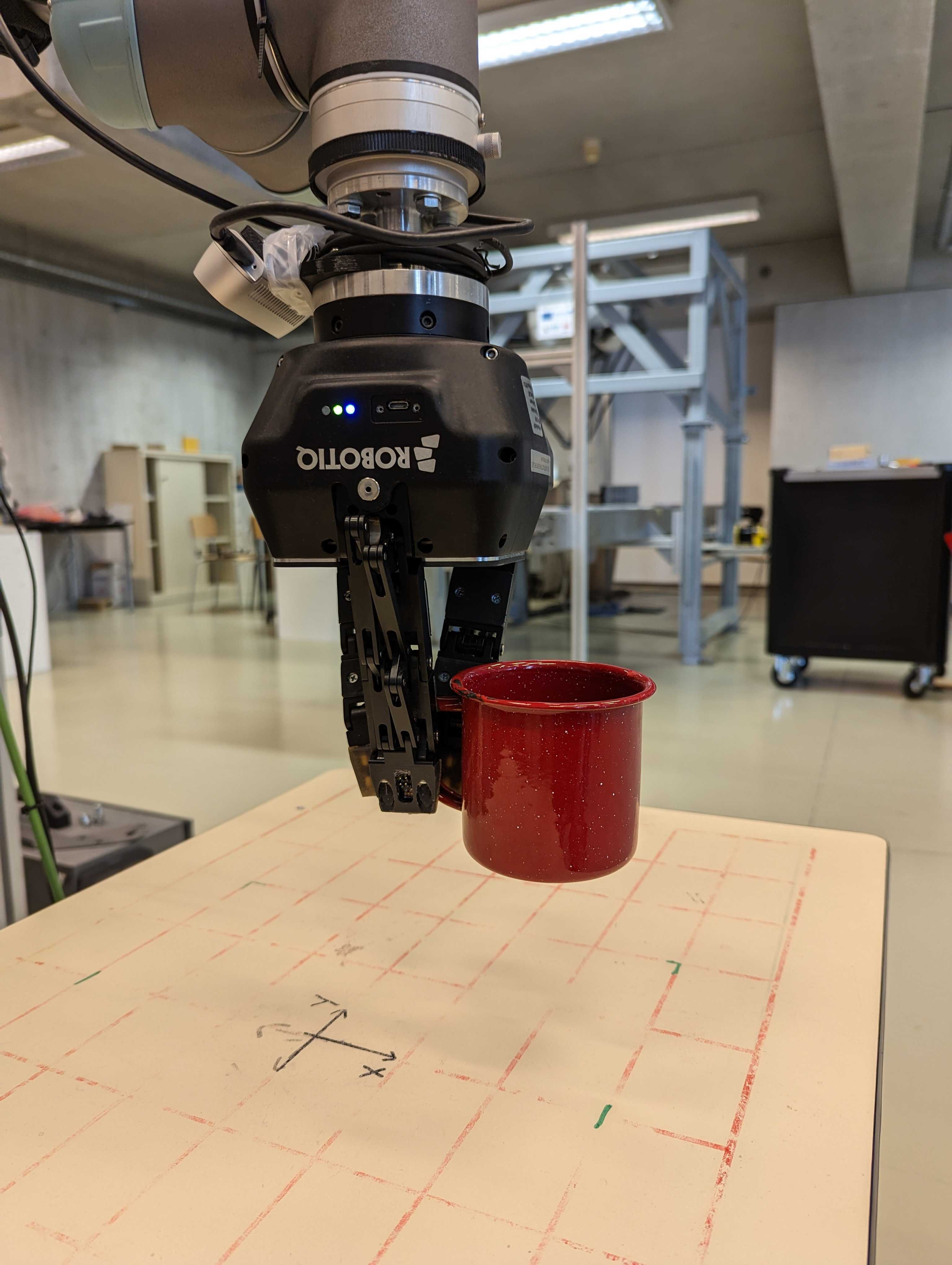}
    }
\caption{(left) Object assets used in the real setup. (right) Successful grasp of a mug.}
    \label{fig:object_assets}
\end{figure}

We adopt the setup from~\cite{marlier2023simulation}. We perform $20$ rounds with $5$ objects among $13$ (see Fig.~\ref{fig:object_assets}), using a protocol similar to the simulation experiments. The objects are chosen based on their availability in the lab and whether they were seen or unseen during training. For novel objects, we achieve a success rate of $\mathbf{95.6}\%$ and a percent cleared of $\mathbf{88}\%$, showing the strong adaptability and performance of our approach. The discrepancy between the simulation and real-world setup is overcome without any decrease in performance. In both simulation and real-world settings, the majority of failure cases are due to insufficient friction forces, causing the objects to slip.

\begin{table}
\caption{Success rates (\%) and \% cleared for picking experiments for the packed scenario with $5$ objects over $200$ rounds.}
\begin{center}
\begin{tabular}{l|c|c}
\hline
\hline
Method & Success rate & \% cleared \\
\hline
{\it Simulation results} & & \\
\,\, GPD \cite{ten2017grasp} & $73.7$ & $72.8$\\
\,\, VGN ($\varepsilon=0.95$) \cite{breyer2020volumetric} & $91.5$ & $79$ \\
\,\, VGN ($\varepsilon=0.9$) \cite{breyer2020volumetric} & $87.6$ & $80.4$ \\
\,\, VGN ($\varepsilon=0.85$) \cite{breyer2020volumetric} & $80.4$ & $79.9$ \\
\,\, Ours & $ 91.1$& $77$ \\
\hline
{\it Real-world results} & & \\
\,\, Ours & $95.6$ & $88$ \\
\hline
\hline
\end{tabular}
\label{tab:results}
\end{center}
\end{table}

\section{Related Work}

Grasp sampling strategies that generate data for machine learning methods can be categorized based on their coverage of the hand configuration space $\mathcal{H}$~\cite{eppner2022billion}. Simple strategies like uniform sampling provide direct density estimation but are highly inefficient. Heuristic methods, on the other hand, mainly rely on object geometry. Notably, the most efficient strategies~\cite{mahler2017dexnet, yan2018learning} are not suitable for complex settings such as multi-fingered grippers. Our occupancy networks-based approach offers direct density estimation, efficient sampling, and does not depend on specific object or gripper assumptions.

Representing a 3D scene as a parameterized function using neural networks has recently gained substantial interest. Occupancy networks~\cite{mescheder2019occupancy} determine whether a point is occupied or not. However, they lack \textit{equivariance} for translations and rotations. To overcome this limitation, translation-equivariant feature planes are derived from convolutional networks, resulting in Convolutional Occupancy Networks~\cite{peng2020convolutional}. Achieving rotation-equivariance is more challenging, but innovative solutions have recently emerged~\cite{chen20223d, deng2021vector}.

Various methods exist for sampling from a distribution defined on a Riemannian manifold. Adapting Markov Chain Monte Carlo methods \cite{liua2022geometry} to Riemannian manifolds allows sampling from a differentiable and tractable likelihood. Our approach eliminates the need for an explicit likelihood. Normalizing flows \cite{rezende2020normalizing} which target density defined on manifolds can rapidly sample from the posterior distribution. However, it remains unclear how to use their gradients for Riemannian optimization.

Finally, probabilistic approaches for grasping problems typically depend on explicit likelihood functions that model the probability of success or a grasp quality metric related to an observation and a grasp pose~\cite{lu2020planning, cai2022real, breyer2020volumetric, van2020learning}. Closer to our work, a similar study~\cite{marlier2023simulation} uses simulation-based inference to compute the maximum a posteriori through Riemannian gradient ascent. However, this approach uses a heuristic prior and cannot accommodate multiple objects.

\section{Conclusion}

We have shown that Bayesian inference can be effectively applied to robotic grasping in complex and noisy environments. Through innovative enhancements, we have improved the sample efficiency of the inference pipeline. Our approach can manage tasks with escalating complexity and proves valuable for real-world robotic applications. Future research will focus on controlling the complete 6-DoF of the robotic hand.

\section*{Acknowledgement}
Norman Marlier acknowledges the Belgian Fund for Research training in Industry and Agriculture for its financial support (FRIA grant). Computational resources have been provided by the Consortium des Équipements de Calcul Intensif (CÉCI), funded by the Fonds de la Recherche Scientifique de Belgique (F.R.S.-FNRS) under Grant No. 2.5020.11 and by the Walloon Region.

\bibliographystyle{IEEEtran}
\bibliography{IEEEabrv, bibliography.bib}

\begin{thebibliography}{10}
\providecommand{\url}[1]{#1}
\csname url@rmstyle\endcsname
\providecommand{\newblock}{\relax}
\providecommand{\bibinfo}[2]{#2}
\providecommand\BIBentrySTDinterwordspacing{\spaceskip=0pt\relax}
\providecommand\BIBentryALTinterwordstretchfactor{4}
\providecommand\BIBentryALTinterwordspacing{\spaceskip=\fontdimen2\font plus
\BIBentryALTinterwordstretchfactor\fontdimen3\font minus
  \fontdimen4\font\relax}
\providecommand\BIBforeignlanguage[2]{{%
\expandafter\ifx\csname l@#1\endcsname\relax
\typeout{** WARNING: IEEEtran.bst: No hyphenation pattern has been}%
\typeout{** loaded for the language `#1'. Using the pattern for}%
\typeout{** the default language instead.}%
\else
\language=\csname l@#1\endcsname
\fi
#2}}

\bibitem{cranmer2020frontier}
K.~Cranmer, J.~Brehmer, and G.~Louppe, ``The frontier of simulation-based
  inference,'' \emph{Proceedings of the National Academy of Sciences}, 2020.

\bibitem{byrne2013geodesic}
S.~Byrne and M.~Girolami, ``Geodesic monte carlo on embedded manifolds,''
  \emph{Scandinavian Journal of Statistics}, vol.~40, no.~4, pp. 825--845,
  2013.

\bibitem{peng2020convolutional}
S.~Peng, M.~Niemeyer, L.~Mescheder, M.~Pollefeys, and A.~Geiger,
  ``Convolutional occupancy networks,'' in \emph{Computer Vision--ECCV 2020:
  16th European Conference, Glasgow, UK, August 23--28, 2020, Proceedings, Part
  III 16}.\hskip 1em plus 0.5em minus 0.4em\relax Springer, 2020, pp. 523--540.

\bibitem{ronneberger2015unet}
O.~Ronneberger, P.~Fischer, and T.~Brox, ``U-net: Convolutional networks for
  biomedical image segmentation,'' 2015.

\bibitem{pmlr-v119-hermans20a}
\BIBentryALTinterwordspacing
J.~Hermans, V.~Begy, and G.~Louppe, ``Likelihood-free {MCMC} with amortized
  approximate ratio estimators,'' in \emph{Proceedings of the 37th
  International Conference on Machine Learning}, ser. Proceedings of Machine
  Learning Research, H.~D. III and A.~Singh, Eds., vol. 119.\hskip 1em plus
  0.5em minus 0.4em\relax PMLR, 13--18 Jul 2020, pp. 4239--4248. [Online].
  Available: \url{http://proceedings.mlr.press/v119/hermans20a.html}
\BIBentrySTDinterwordspacing

\bibitem{murphy2021implicit}
K.~Murphy, C.~Esteves, V.~Jampani, S.~Ramalingam, and A.~Makadia,
  ``Implicit-pdf: Non-parametric representation of probability distributions on
  the rotation manifold,'' \emph{arXiv preprint arXiv:2106.05965}, 2021.

\bibitem{breyer2020volumetric}
M.~Breyer, J.~J. Chung, L.~Ott, S.~Roland, and N.~Juan, ``Volumetric grasping
  network: Real-time 6 dof grasp detection in clutter,'' in \emph{Conference on
  Robot Learning}, 2020.

\bibitem{marlier2023simulation}
N.~Marlier, O.~Br{\"u}ls, and G.~Louppe, ``Simulation-based bayesian inference
  for robotic grasping,'' \emph{arXiv preprint arXiv:2303.05873}, 2023.

\bibitem{ten2017grasp}
A.~Ten~Pas, M.~Gualtieri, K.~Saenko, and R.~Platt, ``Grasp pose detection in
  point clouds,'' \emph{The International Journal of Robotics Research},
  vol.~36, no. 13-14, pp. 1455--1473, 2017.

\bibitem{eppner2022billion}
C.~Eppner, A.~Mousavian, and D.~Fox, ``A billion ways to grasp: An evaluation
  of grasp sampling schemes on a dense, physics-based grasp data set,'' in
  \emph{Robotics Research: The 19th International Symposium ISRR}.\hskip 1em
  plus 0.5em minus 0.4em\relax Springer, 2022, pp. 890--905.

\bibitem{mahler2017dexnet}
J.~Mahler, J.~Liang, S.~Niyaz, M.~Laskey, R.~Doan, X.~Liu, J.~A. Ojea, and
  K.~Goldberg, ``Dex-net 2.0: Deep learning to plan robust grasps with
  synthetic point clouds and analytic grasp metrics,'' 2017.

\bibitem{yan2018learning}
X.~Yan, J.~Hsu, M.~Khansari, Y.~Bai, A.~Pathak, A.~Gupta, J.~Davidson, and
  H.~Lee, ``Learning 6-dof grasping interaction via deep geometry-aware 3d
  representations,'' 2018.

\bibitem{mescheder2019occupancy}
L.~Mescheder, M.~Oechsle, M.~Niemeyer, S.~Nowozin, and A.~Geiger, ``Occupancy
  networks: Learning 3d reconstruction in function space,'' in
  \emph{Proceedings of the IEEE/CVF conference on computer vision and pattern
  recognition}, 2019, pp. 4460--4470.

\bibitem{chen20223d}
Y.~Chen, B.~Fernando, H.~Bilen, M.~Nie{\ss}ner, and E.~Gavves, ``3d equivariant
  graph implicit functions,'' in \emph{Computer Vision--ECCV 2022: 17th
  European Conference, Tel Aviv, Israel, October 23--27, 2022, Proceedings,
  Part III}.\hskip 1em plus 0.5em minus 0.4em\relax Springer, 2022, pp.
  485--502.

\bibitem{deng2021vector}
C.~Deng, O.~Litany, Y.~Duan, A.~Poulenard, A.~Tagliasacchi, and L.~J. Guibas,
  ``Vector neurons: A general framework for so (3)-equivariant networks,'' in
  \emph{Proceedings of the IEEE/CVF International Conference on Computer
  Vision}, 2021, pp. 12\,200--12\,209.

\bibitem{liua2022geometry}
C.~Liua and J.~Zhub, ``Geometry in sampling methods: A review on manifold mcmc
  and particle-based variational inference methods,'' \emph{Advancements in
  Bayesian Methods and Implementations}, vol.~47, p. 239, 2022.

\bibitem{rezende2020normalizing}
D.~J. Rezende, G.~Papamakarios, S.~Racaniere, M.~Albergo, G.~Kanwar,
  P.~Shanahan, and K.~Cranmer, ``Normalizing flows on tori and spheres,'' in
  \emph{International Conference on Machine Learning}.\hskip 1em plus 0.5em
  minus 0.4em\relax PMLR, 2020, pp. 8083--8092.

\bibitem{lu2020planning}
Q.~Lu, K.~Chenna, B.~Sundaralingam, and T.~Hermans, ``Planning multi-fingered
  grasps as probabilistic inference in a learned deep network,'' in
  \emph{Robotics Research}.\hskip 1em plus 0.5em minus 0.4em\relax Springer,
  2020, pp. 455--472.

\bibitem{cai2022real}
J.~Cai, J.~Cen, H.~Wang, and M.~Y. Wang, ``Real-time collision-free grasp pose
  detection with geometry-aware refinement using high-resolution volume,''
  \emph{IEEE Robotics and Automation Letters}, vol.~7, no.~2, pp. 1888--1895,
  2022.

\bibitem{van2020learning}
M.~Van~der Merwe, Q.~Lu, B.~Sundaralingam, M.~Matak, and T.~Hermans, ``Learning
  continuous 3d reconstructions for geometrically aware grasping,'' in
  \emph{2020 IEEE International Conference on Robotics and Automation
  (ICRA)}.\hskip 1em plus 0.5em minus 0.4em\relax IEEE, 2020, pp.
  11\,516--11\,522.

\end{thebibliography}

\onecolumn

\newpage

\appendix

\subsection{Likelihood-free geodesic Monte Carlo}
\label{appendix:algorithm}
\IncMargin{1em}
\begin{algorithm}
\DontPrintSemicolon
\SetAlgoNoLine
\KwIn{A Manifold $\mathcal{M}$
\newline Initial parameters $\mathbf{h}_{0}$
\newline Prior $p(\mathbf{h}\mid \mathbf{P})$
\newline Momentum distribution $p(\mathbf{v})$
\newline Trained classifier $d_{\phi}(S, \mathbf{h}, \mathbf{P})$
\newline Observations $S, \mathbf{P}$}
\KwOut{Markov chain $\mathbf{h}_{1:T}$}
\BlankLine
 $t \leftarrow 0$ \;
 $\mathbf{h}_{t} \leftarrow \mathbf{h}_{0}$ \;
 \For{t \textless T}
 {
    $\mathbf{v}_{t} \sim p(\mathbf{v})$ \;
    $\mathbf{v}_{t} \leftarrow \bm{\pi}_{\mathbf{h}_{t}}(\mathbf{v}_{t})$ \;
    $k \leftarrow 0$ \;
    $\mathbf{v}_{k} \leftarrow \mathbf{v}_{t}$ \;
    $\mathbf{h}_{k} \leftarrow \mathbf{h}_{t}$ \;
    \For{k \textless L}
    {
        $\mathbf{v}_{k} \leftarrow \mathbf{v}_{k} + \frac{\epsilon}{2}\nabla_{\mathbf{h}_{k}}\log r(S\mid \mathbf{h}_{k}, \mathbf{P})$ \;
        $\mathbf{v}_{k} \leftarrow \bm{\pi}_{\mathbf{h}_{k}}(\mathbf{v}_{k})$ \;
        $\mathbf{h}_{k} \leftarrow \gamma(\epsilon), \gamma(0)=\mathbf{h}_{k}$ \;
        $\mathbf{v}_{k} \leftarrow \dot{\gamma}(\epsilon), \dot{\gamma}(0) = \mathbf{v}_{k}$ \;
        $\mathbf{v}_{k} \leftarrow \mathbf{v}_{k} + \frac{\epsilon}{2}\nabla_{\mathbf{h}_{k}}\log r(S\mid \mathbf{h}_{k}, \mathbf{P})$ \;
        $\mathbf{v}_{k} \leftarrow \bm{\pi}_{\mathbf{h}_{k}}(\mathbf{v}_{k})$ \;
        $k \leftarrow k+1$ \;
    }
    $\lambda_{k} \leftarrow \log r(S\mid \mathbf{h}_{k}, \mathbf{P}) + \log p(\mathbf{h}_{k}\mid \mathbf{P}) - \frac{1}{2}\mathbf{v}_{k}^{T}\mathbf{v}_{k} $ \;
    $\lambda_{t} \leftarrow \log r(S\mid \mathbf{h}_{t}, \mathbf{P}) + \log p(\mathbf{h}_{t}\mid \mathbf{P}) - \frac{1}{2}\mathbf{v}_{t}^{T}\mathbf{v}_{t} $ \;
    $\rho \leftarrow \min(\exp(\lambda_{k}-\lambda_{t}), 1)$ \;
    $ \mathbf{h}_{t+1} \leftarrow \begin{cases}
      \mathbf{h}_{k} & \text{with a probability } \rho\\
      \mathbf{h}_{t} & \text{with a probability } 1-\rho\\
    \end{cases}      $\;
    $t \leftarrow t+1$ \;
 }
 \Return{$\mathbf{h}_{1:T}$}
 \caption{Likelihood-free geodesic Hamiltonian Monte Carlo}
 \label{alg:gHMCNRE}
\end{algorithm}
\DecMargin{1em}

\newpage

\subsection{Sampling the orientation: toy problem}
\label{appendix:sampling_orientation}
\begin{figure}[ht]
    \centering
    \includegraphics[scale=0.65]{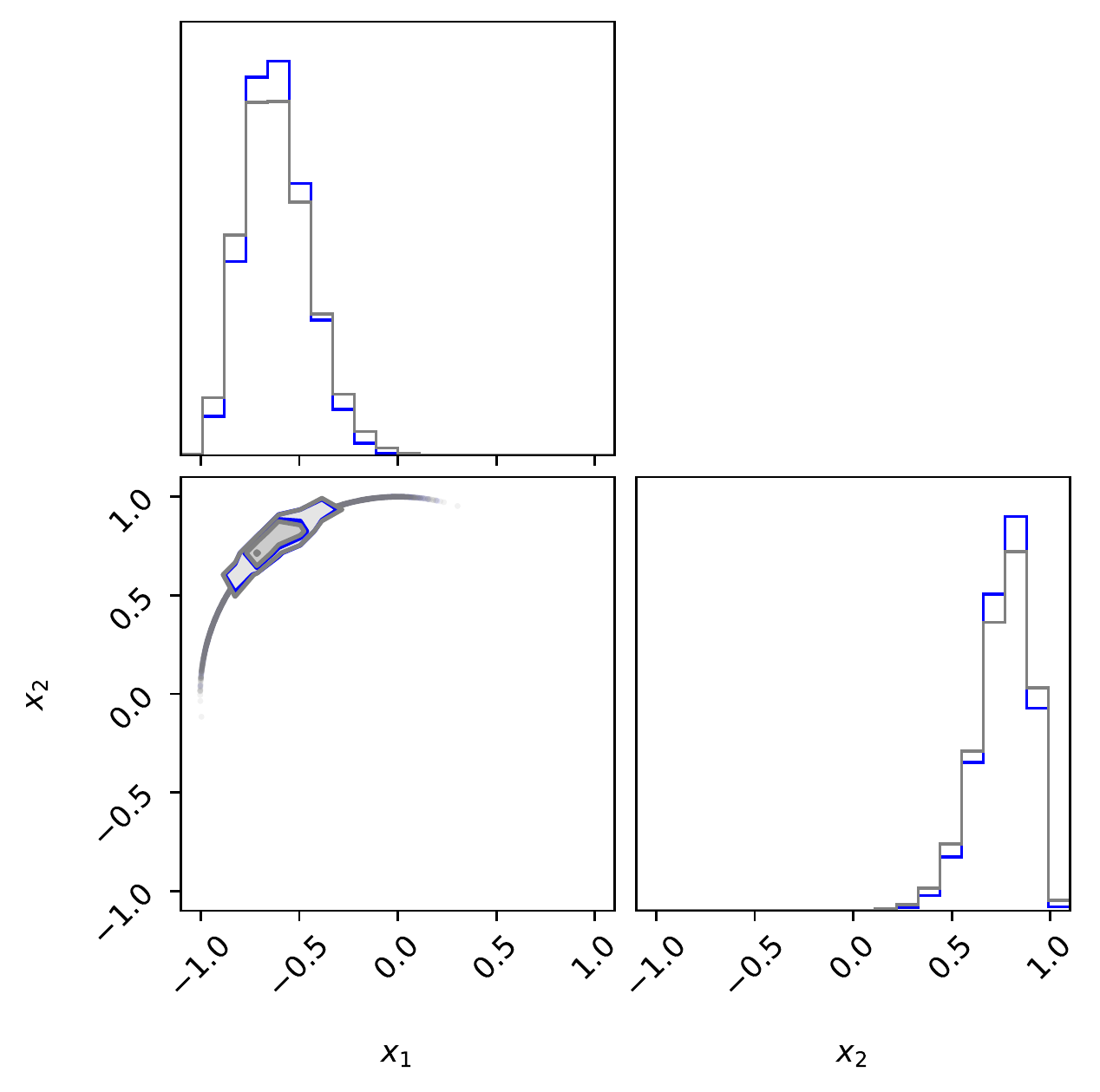}
    
    \caption{Posterior for $\mathbb{S}^{1}$. In blue, the distribution obtained through GMC with the ratio and in grey, the ground truth posterior.}
    \label{fig:toy_problem_sphere}
\end{figure}

Given a model parameter sample $q_{\theta} \in \mathbb{S}^{d}$, the forward generative process is defined as:
\begin{align}
    \nu &= q_{\theta}\\
    \kappa &= 20 \\
    q_{x} &\sim \exp(\kappa\nu^{T}q_{x})
\end{align}
with the prior $p(q_{\theta}) \stackrel{\Delta}{=}\texttt{SphericalUniform}(d)$. It follows the true posterior $p(q_{\theta}\mid q_{x}) \propto \exp(\kappa q_{x}^{T}q_{\theta})$. We use a MLP of 3 layers with 64 neurons to approximate the likelihood-to-evidence ratio. All the activation functions are ReLU except the last one which is linear. We train the ratio with 1000000 samples with a batch size of 8000 for 50 epochs. For the geodesic HMC, we use 100 chains and 2000 transitions with a burn in of 1000. The integration parameters are $\epsilon=0.01, L=20$. The approximate posterior shares the same structure that the true posterior, demonstrating its accuracy (Fig.~\ref{fig:toy_problem_sphere}). Moreover, we conduct a quantitative analysis by computing the Maximal Mean Discrepancy (MMD) for $\mathbb{S}^{1}$ and $\mathbb{S}^{3}$. We obtain respectively $0.0028 \pm 5.84e^{-6}$ and $0.01 \pm 3e^{-5}$ for an identity kernel between the two geodesic means for 10 different observations $q_{x}$.

\newpage

\subsection{Sampling the position: multiple objects scene}
\label{appendix:sampling_position}

\begin{figure}[ht]
    \centering
    \includegraphics[scale=0.6]{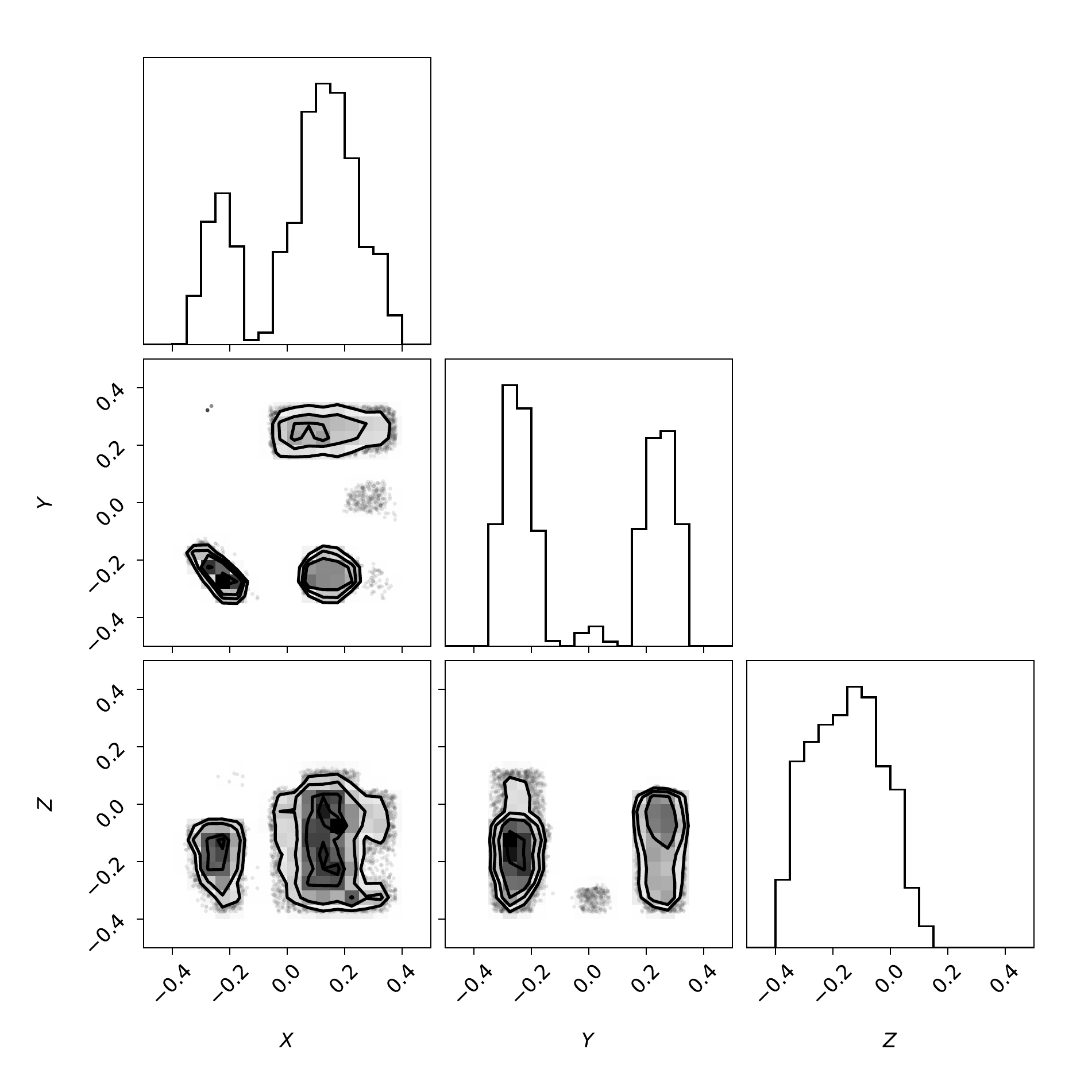}
    \caption{Approximate posterior $p(\mathbf{x} | o=1, \mathbf{P})$ of a scene with 5 different objects.}
    \label{fig:sampling_pos}
\end{figure}

In this experiment, we have a scene containing five objects with varying levels of difficulty, as shown in Fig.~\ref{fig:pipeline}. We use a convolutional occupancy network \cite{peng2020convolutional} that is trained for 120,000 iterations with a batch size of 32 samples. The network has a resolution of 128 and a feature dimension of 32. 
To sample from the posterior  $p(\mathbf{x} | o=1, \mathbf{P}) \propto p(o=1 | \mathbf{x}, \mathbf{P})  p(\mathbf{x})$, as we previously explained, we employed HMC with specific hyper-parameters. These include 100 chains of 5000 transitions and a burn-in of 1000. The
integration parameters are $\epsilon$  = 0.01, $L$ = 20. The chains' initials point are sampled uniformly in the bounding box of the objects. The corner plot in Fig.~\ref{fig:sampling_pos} illustrates the resulting posterior, which aligns with the location and shape of four out of the five objects and the potential grasping point. However, the occupancy of the fifth object cannot be recovered due to either being regarded as noise or having a much smaller density compared to the other objects.

\end{document}